%% file: main.tex
\documentclass[pmlr]{jmlr}

\usepackage{amsmath,amssymb,graphicx,url,algorithm2e}

\makeatletter
\def\set@curr@file#1{\def\@curr@file{#1}} 
\makeatother

\input{macros}
\input{defns}

\jmlrvolume{126}
\jmlryear{2021}
\jmlrworkshop{Machine Learning for Healthcare}

\eat{
\usepackage[
  style=alphabetic,
  citestyle=alphabetic,
  natbib=true,
  backend=bibtex,
  maxcitenames=3,  mincitenames=1,
  maxbibnames=6,  minbibnames=1,
  firstinits=true,
  backref=true, 
  hyperref=true,
  doi=false,  isbn=false,  url=true,
  arxiv=abs
]{biblatex}

}

\usepackage[capitalize]{cleveref}
\Crefname{section}{Sec.}{Secs.}

\usepackage[font={small,it}]{caption}

\usepackage{authblk}

\title[Risk score learning for Covid]
{Risk score learning for COVID-19 contact tracing apps}

\author[1]{Kevin Murphy}
\author[1]{Abhishek Kumar}
\author[1]{Stylianos Serghiou}
\affil[1]{Google Research}

\eat{
\author{\Name{Firstname Lastname}
       \Email{name@email.edu}\\ 
       \addr Department of ML and Health Research\\
       University\\
       City, State, Country 
       \AND
       \Name{Firstname Lastname}
       \Email{name@email.edu}\\ 
       \addr Department of ML and Health Research\\
       University\\
       City, State, Country} 
       }
       
\date{\today}

\begin{document}

\maketitle

\input{abstract}

\input{intro}
\input{simulator}

\input{risk_model}

\input{learning}

\input{related}

\input{discussion}

\small

\bibliography{references}

\end{document}

%% file: macros.tex
\newcommand{\eat}[1]{}

\mathchardef\mhyphen="2D 
\mathchardef\mdash="2D 

\newcommand{\ie}{\emph{i.e.}}

\newcommand{\eg}{\emph{e.g.}}

\newcommand{\myvec}[1]{\mathbf{#1}}

\makeatletter
\newcommand{\oset}[3][-0.3ex]{%
  \mathrel{\mathop{#3}\limits^{
    \vbox to#1{\kern-2\ex@
    \hbox{$\scriptstyle#2$}\vss}}}}
\makeatother

\newcommand{\vp}{\myvec{p}}

\newcommand{\vs}{\myvec{s}}

\newcommand{\vy}{\myvec{y}}

\newcommand{\vU}{\myvec{U}}
\newcommand{\vV}{\myvec{V}}

\newcommand{\mymathcal}[1]{\mathcal{#1}}

\newcommand{\calL}{\mymathcal{L}}

\newcommand{\ind}[1]{\mathbb{I}\left({#1}\right)}

\newcommand{\expect}[1]{\mathbb{E}\left[{#1}\right]} 

\newcommand{\loss}{\calL}

\newcommand{\be}{\begin{equation}}
\newcommand{\ee}{\end{equation}}
\newcommand{\bea}{\begin{eqnarray}}
\newcommand{\eea}{\end{eqnarray}}
\newcommand{\beaa}{\begin{eqnarray*}}
\newcommand{\eeaa}{\end{eqnarray*}}
\newcommand{\ba}{\begin{align*}}
\newcommand{\ea}{\end{align*}}

\newcommand{\figdir}{figures}

\newcommand{\addfig}[4] 
{
    \begin{figure}
    \centering
    \includegraphics[#1]{\figdir/#4}
    \caption{#2}
    \label{fig:#3}
    \end{figure}
}

\newcommand{\addtwofigs}[5]    
{
    \begin{figure}
    \centering
    \begin{subfigure}{0.45\textwidth}
      \centering
      \includegraphics[#1]{\figdir/#4}
      \caption{ }
     \end{subfigure}
~
    \begin{subfigure}{0.45\textwidth}
            \centering
            \includegraphics[#1]{\figdir/#5}
                  \caption{ }
    \end{subfigure}
    \caption{#2}
    \label{fig:#3}
    \end{figure}
}

\newcommand{\addthreefigs}[6]    
{
    \begin{figure}
    \centering
    \begin{subfigure}{0.3\textwidth}
      \centering
      \includegraphics[#1]{\figdir/#4}
      \caption{ }
     \end{subfigure}
~
    \begin{subfigure}{0.3\textwidth}
            \centering
            \includegraphics[#1]{\figdir/#5}
                  \caption{ }
  \end{subfigure}
~
    \begin{subfigure}{0.3\textwidth}
            \centering
            \includegraphics[#1]{\figdir/#6}
                  \caption{ }
    \end{subfigure}
    \caption{#2}
    \label{fig:#3}
    \end{figure}
}

\newcommand{\addfourfigs}[7]    
{
    \begin{figure}
    \centering
    \begin{subfigure}{0.45\textwidth}
      \centering
      \includegraphics[#1]{\figdir/#4}
      \caption{ }
     \end{subfigure}
~
    \begin{subfigure}{0.45\textwidth}
      \centering
      \includegraphics[#1]{\figdir/#5}
      \caption{ }
     \end{subfigure}
\\
    \begin{subfigure}{0.45\textwidth}
      \centering
      \includegraphics[#1]{\figdir/#6}
      \caption{ }
     \end{subfigure}
~
    \begin{subfigure}{0.45\textwidth}
      \centering
      \includegraphics[#1]{\figdir/#7}
      \caption{ }
     \end{subfigure}
    \caption{#2}
    \label{fig:#3}
    \end{figure}
}

\newcommand{\addsixfigs}[9]    
{
    \begin{figure}
    \centering
    \begin{subfigure}{0.45\textwidth}
      \centering
      \includegraphics[#1]{\figdir/#4}
      \caption{ }
     \end{subfigure}
~
    \begin{subfigure}{0.45\textwidth}
      \centering
      \includegraphics[#1]{\figdir/#5}
      \caption{ }
     \end{subfigure}
\\
    \begin{subfigure}{0.45\textwidth}
      \centering
      \includegraphics[#1]{\figdir/#6}
      \caption{ }
     \end{subfigure}
~
    \begin{subfigure}{0.45\textwidth}
      \centering
      \includegraphics[#1]{\figdir/#7}
      \caption{ }
     \end{subfigure}
~
\\
    \begin{subfigure}{0.45\textwidth}
      \centering
      \includegraphics[#1]{\figdir/#8}
      \caption{ }
     \end{subfigure}
~
    \begin{subfigure}{0.45\textwidth}
      \centering
      \includegraphics[#1]{\figdir/#9}
      \caption{ }
    \end{subfigure}
    \caption{#2}
    \label{fig:#3}
    \end{figure}
}

\makeatletter
\newcommand{\chapterauthor}[1]{%
  {\parindent0pt\vspace*{-25pt}%
  \linespread{1.1}\large\scshape#1%
  \par\nobreak\vspace*{35pt}}
  \@afterheading%
}
\makeatother





%% file: defns.tex
\renewcommand{\figdir}{Figures}

\newcommand{\COVID}{COVID-19\xspace}

\newcommand{\GAEN}{GAEN\xspace}

\newcommand{\UKapp}{England/Wales GAEN app\xspace}

\newcommand{\appParams}{\psi}

\newcommand{\simParams}{\phi}
\newcommand{\modelParams}{\simParams}
\newcommand{\modelParam}{\modelParams}
\newcommand{\paramsSim}{\simParams}
\newcommand{\paramSim}{\simParams}

\newcommand{\residBle}{\Delta^{\text{ble}}}

\newcommand{\bleThresh}{\theta^{\text{ble}}}
\newcommand{\infThresh}{\theta^{\text{con}}}

\newcommand{\bleWeights}{w^{\text{ble}}}
\newcommand{\infWeights}{w^{\text{con}}}
\newcommand{\weightsBle}{\bleWeights}
\newcommand{\weightsInf}{\infWeights}

\newcommand{\transRateModel}{\lambda}
\newcommand{\transRateApp}{\mu}

\newcommand{\Nbuckets}{N_B}
\newcommand{\level}{c}

\newcommand{\Nlevels}{N_C}

\newcommand{\expoSetClean}{\mathcal{X}}
\newcommand{\expoSetNoisy}{\tilde{\mathcal{X}}}
\newcommand{\expoSetNoisyPos}{\tilde{\mathcal{X}}}
\newcommand{\expoClean}{x}
\newcommand{\expoNoisy}{\tilde{x}}

\newcommand{\atten}{a}
\newcommand{\duration}{\tau}

\newcommand{\distance}{d}
\newcommand{\tost}{\sigma}

\newcommand{\temperature}{t}

\newcommand{\timeExp}{t^{\text{exp}}}
\newcommand{\timeSym}{t^{\text{sym}}}

\newcommand{\frisk}{f_{\text{risk}}}
\newcommand{\flevel}{f_{\text{con}}}
\newcommand{\fdist}{f_{\text{dist}}}
\newcommand{\finf}{f_{\text{inf}}}

\newcommand{\fhazard}{f_{\text{hazard}}}

\newcommand{\fble}{f_{\text{ble}}}

\newcommand{\riskUser}{R}
\newcommand{\riskEvent}{r}
\newcommand{\qUser}{Q}
\newcommand{\qEvent}{q}

\newcommand{\scoreEvents}{\vs}
\newcommand{\scoreUser}{S}
\newcommand{\scoreEvent}{s}
\newcommand{\pUser}{P}
\newcommand{\pEvent}{p}

\newcommand{\yUser}{Y}
\newcommand{\yEvent}{y}

\newcommand{\assMatSim}{\vU}
\newcommand{\assMatApp}{\vV}

\newcommand{\assBitSim}{u}
\newcommand{\assBitApp}{v}

\newcommand{\Nusers}{J}
\newcommand{\Ngrid}{N}

%% file: abstract.tex
\begin{abstract}
Digital contact tracing apps for \COVID,
such as the one developed by Google and Apple,
need to estimate the risk that a user was infected during a particular exposure, in order to decide whether to notify the user to take precautions,
such as entering into quarantine, or requesting a test.
Such risk score models contain numerous parameters that must be set
by the public health authority. In this paper, we show how to automatically learn these parameters from data.
    
Our method needs access to exposure and outcome data. Although this data is already being collected (in an aggregated, privacy-preserving way) by several health authorities, in this paper we limit ourselves to simulated data, so that we can systematically study the different factors that affect the feasibility of the approach. In particular, we show that the parameters become harder to estimate when there is more missing data (e.g., due to infections which were not recorded by the app), and when there is model misspecification. Nevertheless, the learning approach outperforms a strong manually designed baseline.
Furthermore, the learning approach can adapt even when the risk factors of the disease change, e.g., due to the evolution of new variants, or the adoption of vaccines.
\end{abstract}

%% file: intro.tex
\section{Introduction}

Digital contact tracing (DCT) based on mobile phone technology has been proposed as one of many tools to help
combat the spread of \COVID.
Such apps  have been shown to reduce \COVID infections
in simulation studies
(e.g., \citep{Ferretti2020,Abueg2021,Cencetti2020})
and real-world deployments
 \citep{UKGAEN,Salathe2020,Ballouz2020,Masel2021,Rodriguez2021,Kendall2020,Huang2020}.
 For example, in a 3-month period,   \citet{UKGAEN}
 estimated that the app used in England and Wales led to about 284,000--594,000 averted infections and 4,200--8,700 averted deaths.
 
DCT apps work by notifying the app user if they have had a "risky encounter" with an index case (someone who has been diagnosed as having \COVID at the time of the encounter). Once  notified, the user may then be advised by their Public Health Authority (PHA) to enter into quarantine, just as they would if they had been contacted by a manual contact tracer. The key question which we focus on in this paper is: 
how can we estimate the probability than an exposure resulted in an
infection? If the app can estimate this reliably, 
it  can decide who to notify based on 
a desired false positive / false negative rate.

To quantify the risk of an encounter, we need access to some (anonymized) features which characterize the encounter.
In this paper, we focus on the features collected by the
Google/ Apple Exposure Notification (GAEN) system
\citep{GAEN},
although our techniques could be extended to work with other DCT apps.
The GAEN app uses 3 kinds of features to
characterize each exposure: the duration of the encounter, the bluetooth signal strength (a proxy for distance between the two people), and (a quantized version of) the infectiousness of the index case (as estimated by the days since their symptoms started, or the date of their positive test).

Let the  set of observed features for the $n$'th encounter be denoted by $\expoNoisy_n$, and let  $\expoSetNoisyPos_{j} = \{\expoNoisy_n: j_n=j\}$ be the set of all exposure features for user $j$.
Finally, let $\yUser_j=1$ if user $j$ gets infected from one (or more) of these encounters,
and $\yUser_j=0$ otherwise. The primary goal of the risk score model is to estimate
the probability of infection, $p(\yUser_j=1|\expoSetNoisyPos_{j};\appParams)$.
Although PHAs are free to use any model they like to compute this quantity, the vast majority have adopted the risk score model which we describe in \cref{sec:riskScore}, since this is the one implemented in the Google/Apple app.

A key open problem is how to choose the parameters $\appParams$
of this risk score model.
Although expert guidance for how to set these parameters has been provided (e.g. \citep{LFPH}), in this paper  we ask if we can  do better using a data-driven approach.
In particular, we assume the PHA has access to a set of
anonymized, labeled features $\{\expoSetNoisyPos_{j}, \yUser_j\}$,
where $\yUser_j \in \{0,1\}$ is the test result for user $j$,
and $\expoSetNoisyPos_{j}$ is the set of exposure features recorded by their app.
Given this data, we can try to optimize the parameters of the risk score model $\appParams$  using weakly-supervised machine learning methods, as we explain in \cref{sec:learning}.

Since we do not currently have access to such "real world" labeled datasets,
in this paper, we create a synthetic dataset  using a simple simulator,
which we describe in \cref{sec:sim}.
We then study the ability of the ML algorithm to estimate the
risk score parameters from this simulated data, as we vary 
the amount of missing data and label noise.
We compare our learned model to a widely used manually created baseline, and show that the ML method performs better (at least in our simulations).
We also show that it is more robust to changes in the true distribution than the manual baseline.\footnote{
A Python Jupyter notebook to reproduce the experiments
can be found at \url{https://github.com/google-research/agent-based-epidemic-sim/tree/develop/agent_based_epidemic_sim/learning/MLHC_paper_experiments.ipynb}.
}

\subsection*{Generalizable Insights about Machine Learning in the Context of Healthcare}

In this paper, we show that it is possible to learn interpretable risk score models using standard machine learning tools, such as 
multiple instance learning
and stochastic gradient descent,
provided we have suitable data.
However, some new techniques are also required,
such as replacing hard thresholding with soft binning, and using constrained optimization methods to ensure monotonicity of the learned function.
We believe these methods could be applied to learn other kinds of risk score models.


%% file: simulator.tex
\section{A simple simulator of \COVID transmission}
\label{sec:infModel}
\label{sec:sim}

The probability of infection from an exposure event depends on many factors, including the duration of the exposure, the distance between the index case (transmitter) and the user (receiver),
the infectiousness of the index case, as well as other unmeasured factors,
such as mask wearing, air flow, etc.
In this section, we describe a simple probabilistic model of \COVID transmission,
which only depends on the factors that
are recorded by the Google/Apple app.
This model forms the foundation of the GAEN risk score model
in \cref{sec:riskScore}.
We will also use this model to generate our simulated training and test data,
as we discuss in \cref{sec:learning}.

\subsection{Single exposure}

Let $\duration_n$ be the duration (in minutes) of the $n$'th exposure,
and let 
$\distance_n$ be the distance (in meters) between the two people during this exposure.
(For simplicity, we assume the distance is constant during the entire interval; we will relax this assumption later.)
Finally, let
$\tost_n$ be the time since of the onset of symptoms of the index case
at the time of exposure, i.e., 
$\tost_n = \timeSym_{i_n} - \timeExp_{j_n}$,
where $i_n$ is the index case for this exposure,
$j_n$ is the user, 
$\timeSym_{i_n}$ is the time of symptom onset for $i_n$,
and  $\timeExp_{j_n}$ is the time that $j_n$ gets exposed to $i_n$.
(If the index case did not show symptoms, we use the date that they tested positive, 
and shift it back in time by 7 days, as a crude approximation.)
Note that $\tost_n$ can be negative.
In \citep{Ferretti2020timing}, they show that $\tost_n$ can be used to estimate
the degree of contagiousness of the index case.

Given the above quantities, 
we define the "hazard score" for this exposure as follows:
\begin{align}
\scoreEvent_n &=
 \fhazard(\duration_n, \distance_n, \tost_n; \paramsSim)
 =
  \duration_n  \times \fdist(\distance_n; \paramsSim) 
  \times \finf(\tost_n; \paramsSim)
  \label{eqn:OurHazard}
\end{align}
where
$\duration_n$ is the duration,
$\distance_n$ is the distance,
$\tost_n$ is the time since symptom onset,
$\fdist(\distance_n;\paramSim)$ is
the simulated risk given distance,
$\finf(\tost_n;\paramSim)$ is the simulated
risk given time,
and $\simParams$ are parameters of the simulator
(as opposed to $\appParams$, which are parameters
of the risk score model that the phone app needs
to learn).

The "correct" functional form for the dependence on distance is unknown.
In this paper, we follow \citep{Briers2020},
who model both short-range (droplet) effects,
as well as longer-range (aerosol) effects, using the following simple
truncated quadratic model:
\begin{align}
  \fdist(\distance; \simParams) = \min(1, D^2_{\min}/\distance^2)
  \label{eqn:fdistQuad}
\end{align}
They propose to set $D^2_{\min}=1$ based on
an argument from the physics of \COVID droplets.

The functional form for the dependence on symptom onset is better understood.
In particular, \citep{Ferretti2020timing} consider a variety of models,
and find that the one with the best fit to the empirical data
\eat{
is a Student $t$ distribution:
\begin{align}
    \finf(\tost;\simParams) \propto \text{St}(\tost|\text{shift}=-0.0747,
    \text{scale}=1.8567, \text{dof}=3.3454)
    \label{eqn:finfStudent}
\end{align}
}
was a scaled skewed logistic distribution:
\begin{align}
    \finf(\tost; \simParams) \propto \text{SkewedLogistic}(\tost|
    \mu=-4,
    \sigma = 1.85,
  \alpha = 5.85,
  \tau = 5.42)
    \label{eqn:finfStudent}
\end{align}

In order to convert the hazard score into a probability of infection (so we can generate samples), we use
a standard exponential dose response model \citep{Smieszek2009,Haas2014}:
\begin{align}
p_n &= \Pr(\yEvent_n=1|\expoClean_n; \paramsSim) = 
1-e^{-\transRateModel s_n}
\end{align}
where $\yEvent_n=1$ iff exposure $n$ resulted in an infection,
and $\expoClean_n=(\duration_n,\distance_n,\tost_n)$ are the features
of this exposure.
The parameter $\transRateModel$  is a fixed constant
with value $3.1 \times 10^{-6}$,
chosen to match the empirical attack rate
reported in \citep{Wilson2020}.

\subsection{Multiple exposures}

A user may have multiple exposures.
Following standard practice,
we assume each exposure could independently infect the user,
so
\begin{align}
  \pUser_j &= \Pr(\yUser_j=1|\expoSetClean_j;\simParams) 
  =  1-\prod_{n \in E_j} (1-  \pEvent_{n}) \\
  &=  1-\prod_{n \in E_j} e^{-\transRateModel \scoreEvent_n}= 
  1-e^{-\transRateModel \sum_n \scoreEvent_n} \\
  &= p_{1} + (1-p_{1}) \times p_{2} + 
  (1-p_{1}) (1-p_{2}) \times p_{3} + \cdots
  \end{align}
 where $E_j$ are all of $j$'s exposure events,
 and $\expoSetClean_j$ is the corresponding set of features.
 
To account for possible "background" exposures that were not recorded by the app, we can add a $\pEvent_0$ term, to reflect the prior probability of an exposure for this user (e.g., based on the prevalence).
This model has the form
\begin{align}
  \pUser_j &= 
   1- (1-\pEvent_0) \prod_{n \in E_j} (1-  \pEvent_{n})
   \label{eqn:pUser}
  \end{align}

\subsection{A bipartite social network}
\label{sec:grid}

Suppose we have a set of $J$ users.
The probability that user $j$ gets infected is given by
\cref{eqn:pUser}. However, it remains to specify the set
of exposure events $E_j$ for each user.
To do this, we create a pool of $N$ randomly generated exposure events, uniformly spaced across the 3d grid of
distances (80 points from 0.1m to 5m),
durations (20 points from 5 minutes to 60 minutes),
and
symptom onset times (-10 to 10 days).
This gives us a total of $N=33600$ exposure events.

We assume that each user $j$ is exposed to a random subset of one or more of these events (corresponding to having encounters with different index cases).
In particular, 
let $\assBitSim_{jn}=1$ iff user $j$ is exposed to event $n$.
Hence the probability that $j$ is infected is given by
    $\pUser_j = 1-\exp[-\transRateModel \scoreUser_j]$,
    where
   $\scoreUser_j = \sum_{n=1}^N  \assBitSim_{jn}  \scoreEvent_n$
   and
    $\scoreEvent_n = \fhazard(\expoClean_n; \paramsSim)$,
where $\expoClean_n=(\duration_n, \distance_n, \tost_n)$.
We can write this in a more compact way as follows.
Let $\assMatSim=[\assBitSim_{jn}]$ be an $\Nusers \times \Ngrid$ binary assignment matrix, and $\scoreEvents$ be the vector of event hazard scores. 
For example, suppose there are $\Nusers=3$ users
and $\Ngrid=5$ events, and we have the following
assignments:
user $j=1$ gets exposed to event $n=1$;
user $j=2$ gets exposed to events $n=1$, $n=2$ and $n=3$;
and
user $j=3$ gets exposed to events $n=3$ and $n=5$.
Thus
\begin{align}
    \assMatSim = \begin{pmatrix}
    1 & 0 & 0 & 0 & 0\\
    1 & 1 & 1 & 0 & 0 \\
    0 & 0 & 1 & 0 & 1
    \end{pmatrix}
    \label{eqn:MILmatrix}
\end{align}
The corresponding vector of true infection probabilities,
one per user, is then given by
$\vp = 1-\exp[-\transRateModel \vU \vs]$.
From this, we can sample a vector of infection  labels, $\vy$,
one per user.
If a user gets infected, we assume they are removed from the population, and cannot infect anyone else (i.e., the pool of $N$ index cases is fixed).
This is just for simplicity. We leave investigation of more realistic population-based simulations to future work.

\subsection{Censoring}

In the real world, a user may get exposed to events that are not recorded by their phone.
We therefore  make a distinction between the events that a user was actually exposed to, encoded by
 $\assBitSim_{jn}$,
 and the events that are visible to the app,
 denoted by $\assBitApp_{jn}$.
 We require that $\assBitApp_{jn}=1 \implies \assBitSim_{jn}=1$, but not vice versa.
 For example, \cref{eqn:MILmatrix} might give rise to the following censored assignment matrix:
 \begin{align}
    \assMatApp = \begin{pmatrix}
    1 & 0 & 0 & 0 & 0\\
    1 & 1 & \mathbf{0} & 0 & 0 \\
    0 & 0 & 1 & 0 & \mathbf{0}
    \end{pmatrix}
    \label{eqn:MILmatrixCensored}
\end{align}
where the censored events are shown in bold.
This censored version of the data is what is made available to the user's app when estimating the exposure risk,
as we discuss in \cref{sec:learning}.
We assume censoring happens uniformly at random.
We leave investigation of more realistic censoring simulations to future work.

\subsection{Bluetooth simulator}
\label{sec:bluetooth}
\label{sec:distanceToBluetooth}

DCT apps do not  observe distance directly.
Instead, they estimate it based on  bluetooth attenuation.
In order to simulate the bluetooth signal, which will be fed as input
to the DCT risk score model,  we use the stochastic forwards model from \citep{Lovett2020}.
This models the atttenuation as a function of distance
using a log-normal distribution, with a mean given by
\be
\expect{\atten|\distance;\modelParams}
= e^{\modelParam_{\alpha} + \modelParam_{\beta} \log(\distance)} 
= e^{\modelParam_{\alpha}} \distance^{\modelParam_{\beta}}
\label{eqn:distToRSSI}
\ee
Using empirical data from MIT Lincoln Labs,  
\citet{Lovett2020} estimate the offset to be 
$\modelParam_{\alpha}=3.92$
and the slope to be
$\modelParam_{\beta}=0.21$.

In this paper, we assume this mapping is deterministic,
even though in reality it is quite noisy,
due to multipath reflections and other environmental factors
(see e.g., \citep{Leith2020plos}).
Fortunately, various methods have been developed to try
to "denoise" the bluetooth signal.
For example, the
\UKapp  uses an approach based
on unscented Kalman smoothing
 \citep{Lovett2020}.
We leave the study of the effect of bluetooth
noise on our learning methods to future work.

%% file: risk_model.tex
\section{The app's risk score model}
\label{sec:GAEN}
\label{sec:riskScore}

In this section, we describe risk score model used by Google/Apple Exposure Notification (GAEN) system.
This can be thought of as a simple approximation to the biophysical
model 
we discussed in \cref{sec:infModel}.

For every exposure recorded by the app, 
the following risk score is computed:
\begin{align} 
 \riskEvent_n &= \frisk(\duration_n, \atten_n, \level_n; \appParams)
 = \duration_n \times \fble(\atten_n; \appParams)
 \times \flevel(\level_n; \appParams)
\end{align}
where
$\duration_n$ is the duration,
$\atten_n$ is the bluetooth attenuation,
$\level_n$ is a quantized version of symptom onset time,
and where  $\fble$ and $\flevel$ are functions defined below.

We can convert the risk score   into an estimated probability
of being infected using
\begin{align}
\qEvent_n& =\Pr(\yEvent_n=1|\expoNoisy_n; \appParams) 
 = 1-\exp[-\transRateApp \riskEvent_n]
 \end{align}
where $\expoNoisy_n = (\duration_n, \atten_n, \level_n)$ are the observed features recorded by the GAEN app,
and $\transRateApp$ is a learned scaling parameter,
analogous to $\transRateModel$ for the simulator.

\subsection{Estimated risk vs bluetooth attenuation}
\label{sec:bluetoothRisk}

\begin{figure}
\centering
\includegraphics[height=2in]{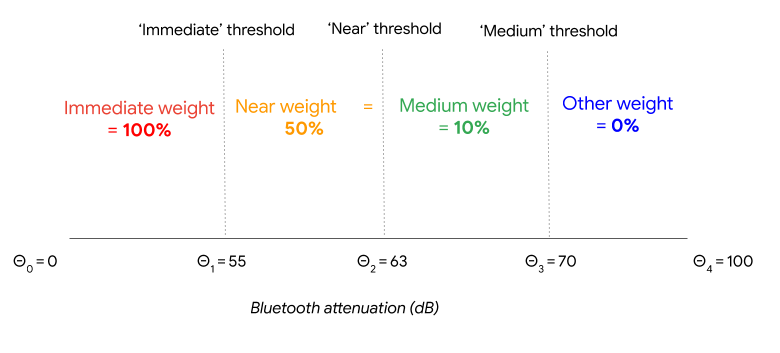}
\caption{3 thresholds defines 4 attenuation buckets, each 
  of which can be given a weight.
  We define $\bleThresh_0=0$ and $\bleThresh_4=100$ as boundary conditions.
   These weights encode the assumption that smaller attenuations (corresponding to closer exposures) have higher risk than larger attenuations.
  }
\label{fig:buckets}
\end{figure}

The GAEN risk score model makes  a piecewise constant approximation
to the risk-vs-attenuation function.
In particular, it defines 3 attenuation thresholds, $\bleThresh_{1:3}$,
which partitions the real line into $4$ intervals.
Each interval or bucket is assigned a weight, $\weightsBle_{1:4}$,
as shown in \cref{fig:buckets}.

Overall, we can view this as approximating
$\fdist(d_n;\simParams)$ by
$\fble(a_n;\appParams)$, where 
\begin{align}
\fble(\atten_n;\appParams)
 &= \begin{cases}
  \weightsBle_1 &\mbox{if $\atten_n \leq \bleThresh_1$} \\
   \weightsBle_2 &\mbox{if $\bleThresh_1 < \atten_n \leq \bleThresh_2$} \\
    \weightsBle_3 &\mbox{if $\bleThresh_2 < \atten_n \leq \bleThresh_3$} \\
     \weightsBle_4 &\mbox{if $\bleThresh_3 > \atten_n$} 
  \end{cases}
\end{align}

\subsection{Estimated risk vs days since symptom onset}
\label{sec:levelRisk}

\begin{figure}[h!]
\centering
\includegraphics[height=2in]{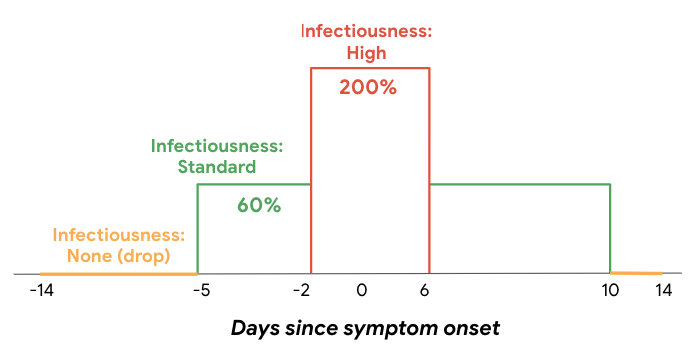}
\caption{A mapping 
  from symptom onset
  to 3 infectiousness levels with corresponding weights.
}
\label{fig:inflevel}
\end{figure}

The health authority can compute the days since symptom
onset for each exposure,
$\tost_n = \timeSym_{i_n} - \timeExp_{j_n}$,
by asking the index case when they first showed symptoms.
(If they did not show any symptoms, they can use a heuristic, such as the date of their test shifted back by several days.)
However, sending the value of $\tost_n$ directly
was considered to be a security risk by Google/Apple.
Therefore, to increase privacy, the symptom onset
value is mapped into  one of 3 infectiousness
or contagiousness levels; this mapping is 
defined by a lookup table
as shown in  \cref{fig:inflevel}.
The three levels  are called "none", "standard" and "high".
We denote this mapping by
\[
\underbrace{\level_{n,l}}_{\text{contagiousness level}}
= \ind{\infThresh_{l-1} < 
\underbrace{\tost_n}_{\text{time since symptom onset}}
\leq \infThresh_l}
\]

Each infectiousness level is associated with a weight.
(However, we require that the weight for
level "none" is $\weightsInf_1=0$, so there are only 2 weights that need to be specified.) Overall, the mapping and the weights define a piecewise constant approximation to the risk vs symptom onset function
defined in \cref{eqn:finfStudent}.
We can write this approximation as
$\finf(\tost_n;\appParams)
\approx \flevel(\text{LUT}(\tost_n);\appParams)$,
where
\begin{align}
\flevel(\level_n;\appParams)
 &= \begin{cases}
 0 &\mbox{if $\level_n=1$} \\
  \weightsInf_2 &\mbox{if $\level_n=2$} \\
    \weightsInf_3 &\mbox{if $\level_n=3$} 
  \end{cases}
\end{align}

\subsection{Comparing the simulated and estimated risk score models}

 \begin{figure}
\centering
\includegraphics[height=3in]{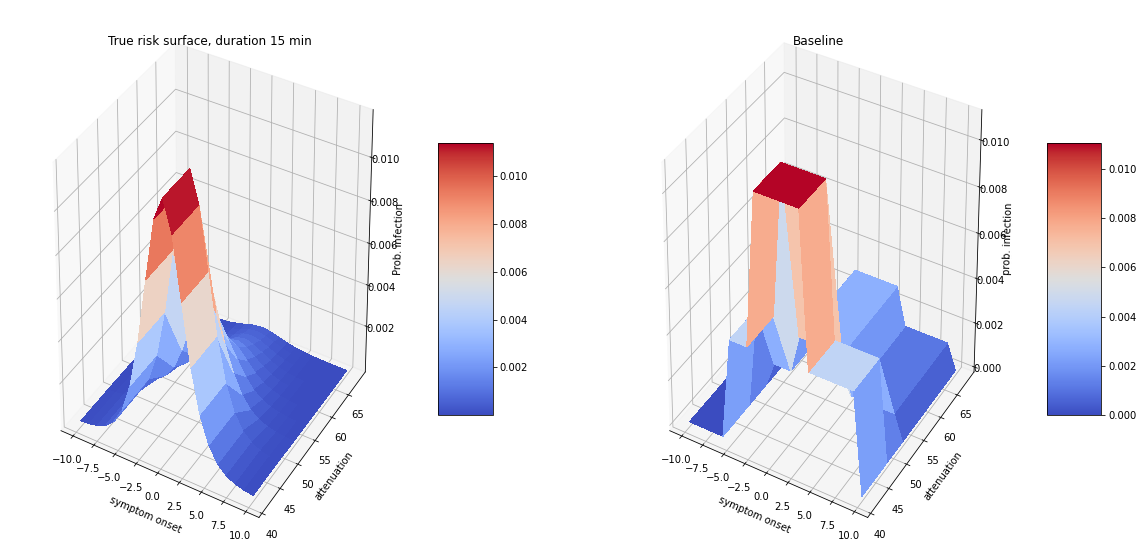}
\caption{
Risk surface used by (left) our simulator and (right) the app.
We plot risk as a function of attenuation and symptom onset.
We assume the duration of the exposure is 
 $\duration=15$ minutes.
  }
\label{fig:riskSurface}
\end{figure}

In \cref{fig:riskSurface}(left),  we plot the 
probability of infection according to our simulator
 as a function of symptom onset $\tost_n$ and distance $\distance_n$,
for a fixed duration $\duration_n=15$.
We see that this is the convolution of the 
(slightly asymmetric)
bell-shaped curve from
the symptom onset function in \cref{eqn:finfStudent}
along the x-axis
with the truncated quadratic curve from
the distance function in 
\cref{eqn:fdistQuad}
along the z-axis.
We can approximate this risk surface using the GAEN piecewise constant approximation, as shown in \cref{fig:riskSurface}(right).

\subsection{Multiple exposures}

We combine the risk from multiple exposure windows by adding the risk scores,
reflecting the assumption of independence between exposure events.
For example,
if there are multiple exposures within an exposure window,
each with different duration and attenuation, we can combine them as follows.
First we compute the total time spent in each bucket:
\begin{align}
\underbrace{\duration_{nb}}_{\text{duration in bucket $b$}} &= 
 \sum_{k=1}^{K_n}
 \underbrace{\duration_{nk}}_{\text{duration}} \times 
\underbrace{\ind{\bleThresh_{b-1} < \atten_{nk} \leq \bleThresh_b}}_{\text{attenuation  is in bucket $b$}}
\label{eqn:bucketing}
\end{align}
where $K_n$ is the number of "micro exposures" within the $n$'th exposure window,
$\duration_{nk}$ is the time spent in the $k$'th micro exposure,
and $\atten_{nk}$ is the corresponding attenuation.
(This gives a piecewise constant approximation to the distance-vs-time curve for any given interaction between two people.)
We then compute the overall risk for this exposure
using the following bilinear function:
\begin{align}
  \underbrace{\riskEvent_n}_{\text{risk score}}  
  &=
  \underbrace{
\left[ \sum_{b=1}^{\Nbuckets} \duration_{nb} \weightsBle_b \right]
}_{\text{weighted exposure minutes}}
\times
\underbrace{
\left[ \sum_{\ell=1}^{\Nlevels} \ind{\level_{n,\ell}} \weightsInf_\ell \right]
}_{\text{weighted contagiousness level}}
\label{eqn:riskScore}
\end{align}
where $\Nbuckets=4$ is the number of attenuation buckets,
and $\Nlevels=3$ is the number of contagiousness levels.

If we have multiple exposures for a user, we sum the risk scores to get
$\riskUser_j  =\sum_{n \in E_j} \riskEvent_n$,
which we can convert to a probability of infection
using $\qUser_j = 1-e^{-\transRateApp \riskUser_j}$.

%% file: learning.tex
\section{Learning the risk score}
\label{sec:learning}

In this section, we discuss how to optimize the parameters of the risk score model
using  machine learning.
We assume access to a labeled dataset, $\{ (\expoSetNoisyPos_j, \yUser_j): j=1:\Nusers\}$,
for a set of $\Nusers$ users. 
This data is generated using the simulator described
in \cref{sec:sim}.

We assume the lookup table mapping  from symptom onset to infectiousness level is fixed,
as shown in \cref{fig:inflevel}; however,
we assume the corresponding infection weights $\weightsInf_2$ and $\weightsInf_3$ are unknown.
Similarly, we assume the 3 attenuation thresholds $\bleThresh_{1:3}$,
as well as the 4 corresponding weights, $\weightsBle_{1:4}$, are unknown.
Finally, we assume the scaling factor $\transRateApp$ is unknown.
Thus there are 10 parameters in total to learn; we denote these by $\appParams=(\weightsInf_{2:3},\bleThresh_{1:3},\weightsBle_{1:4},\transRateApp)$.
It is clear  that the parameters are not uniquely identifiable. For example, we could increase $\transRateApp$ and decrease $\weightsInf$ and the effects would cancel out. Thus there are many parameter settings that all obtain the same maximum likelihood. Our goal is just to identify one element of this set. 

In the sections below,
we describe some of the challenges in learning these parameters, and then our experimental results.

\input{weakly}

\input{optim}
\input{results}

%% file: weakly.tex
\subsection{Learning from data with label noise and censoring}
\label{sec:objective}

We optimize the parameters by maximizing the log-likelihood, or equivalently,
minimizing the binary cross entropy:
\begin{align}
    \loss(\appParams) = -\sum_{j=1}^{\Nusers} \yUser_j \log \qUser_j 
    + (1-\yUser_j) \log (1-\qUser_j)
    \label{eqn:NLL}
\end{align}
where $\yUser_j \in \{0,1\}$ is the infection label for user $j$ coming from the simulator, and
\begin{align}
    \qUser_j &= 1-\exp\left[-\transRateApp \sum_{n \in E_j}
    \frisk(\expoNoisy_n;\appParams) \right]
\end{align}
is the estimated probability of infection, which is computed
using the bilinear risk score model
in \cref{eqn:riskScore}.

A major challenge in optimizing the above objective arises due to the fact that a user may encounter multiple exposure events, but we only see the final outcome from the whole set, not for the individual events.
This problem is known as  multi-instance learning
(see e.g., \citep{Foulds2010}).
For example, consider \cref{eqn:MILmatrix}.
If we focus on user $j=2$, we see that they were exposed
to a "bag" of 3 exposure events (from index cases 1, 2, and 3).
If we observe that this user gets infected
(i.e., $Y_j=1$), we do not know which of these events was the cause.
Thus the algorithm does not know which set of features to pay attention to,
so the larger the bag each user is exposed to, the more challenging
the learning problem.
(Note, however, that if $Y_j=0$, then we know that all the events in the bag must be labeled as negative, since we assume (for simplicity) that the test labels are perfectly reliable.)

In addition, not all exposure events get recorded by the \GAEN app. For example, the index case $i$ may not be using the app, or the exposure may be due to environmental transmission (known as fomites).
This can be viewed as a form of measurement "censoring",
as illustrated in \cref{eqn:MILmatrixCensored},
which is a sparse subset of the true infection matrix in 
 \cref{eqn:MILmatrix}.
This can result in a situation in which 
all the {\em visible} exposure events in the bag are low risk,
but  the user
is infected anyway, because the true cause of the user's infection is not part of $\expoSetNoisy_j$.
This kind of false positive
is a form of label noise which further complicates learning.

%% file: optim.tex
\subsection{Optimization}

In this section, we discuss some algorithmic
issues that arise when trying to optimize the objective in \cref{eqn:NLL}.

\subsubsection{Monotonocity}

We want to ensure that the risk score is monotonically increasing
in attenuation. To do this, 
we can order the  attenuation buckets  from low risk to high, so
$\duration_{n1}$ is time spent in lowest risk bin (largest attenuation), and
 $\duration_{n4}$ is the time spent in the highest risk bin (smallest attenuation).
 Next we  reparameterize $\weightsBle$ as follows:
 \begin{align}
 \weightsBle = [\weightsBle_1,
 \weightsBle_2 = \weightsBle_1 + \residBle_2, 
 \weightsBle_3 = \weightsBle_2 + \residBle_3, 
 \weightsBle_4 = \weightsBle_3 + \residBle_4]
 \end{align}
 Then we optimize over $(\weightsBle_1, \residBle_2, \residBle_3, \residBle_4)$,
 where we 
 use  projected gradient descent to ensure
 $\residBle_b > 0$.
 We can use a similar trick to ensure the risk is monotonically
 increasing with  the infectiousness level.

\subsubsection{Soft thresholding}
\label{sec:softThresholding}

\begin{figure}
\centering
\begin{tabular}{cc}
\includegraphics[height=1.5in]{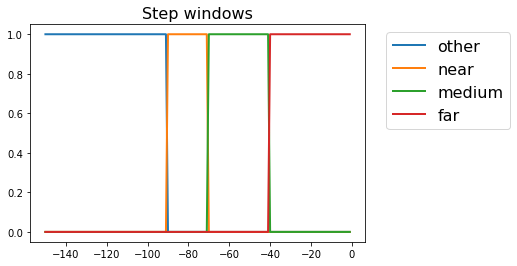}
&
\includegraphics[height=1.5in]{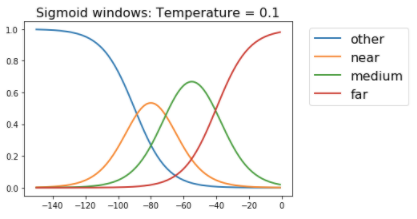}
\\
\includegraphics[height=1.5in]{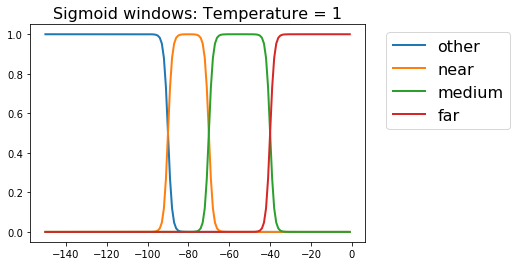}
&
\includegraphics[height=1.5in]{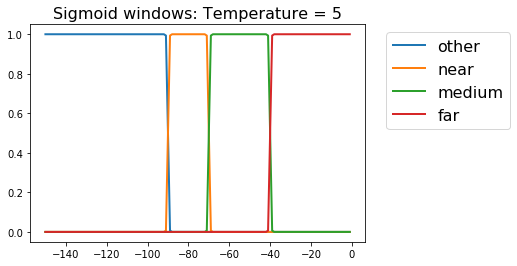}
\end{tabular}
\caption{
(Top left) Hard threshold.
(Other). Sigmoidal approximation at increasing temperature.
}
\label{fig:softThresholding}
\end{figure}

The loss function is not differentiable wrt the attenuation
thresholds $\bleThresh_b$.
We consider two solutions to this.
In the first approach, we use a gradient-free optimizer for
$\bleThresh$
in the outer loop (e.g., grid search), and a gradient-based
optimizer for the weights $\weightsBle_b$ in the inner loop.

In the second approach,
we replace the hard binning in \cref{eqn:bucketing} 
with soft binning, as follows:
\begin{align}
\duration_{nb} &= \sum_k \duration_{nk} 
\; \ind{\bleThresh_{b-1} < \atten_{nk} \leq \bleThresh_b} \\
&\approx \sum_k \duration_{nk} \; 
\sigma_{\tau}(\atten_{nk} - \bleThresh_{b-1})
\sigma_{\tau}(\bleThresh_b - \atten_{nk}) 
 \end{align}
 where
     $\sigma_{\temperature}(x) = \frac{1}{1+e^{-\temperature x}}$
 is the sigmoid (logistic) function, and
 $\temperature$ is a temperature parameter.
The effect of $\temperature$ is illustrated in \cref{fig:softThresholding}.
When learning, we start with a small temperature, 
and then gradually increase it until we approximate the hard-threshold
form, 
as required by the \GAEN app.
 
 The advantage of the soft binning approach
 is that we can optimize the thresholds, as well as the weights, using gradient descent.
 This is much faster and simpler than the nested optimization approach, in which we use grid search in the outer loop.
 Furthermore,
 preliminary experiments suggested that the two approaches yield similar results. We will therefore focus on soft binning for the rest of this paper.
 

%% file: results.tex
\subsection{Experiments}
\label{sec:results}

In this section, we describe our experimental results.

\subsubsection{Setup}

\begin{figure}
    \centering
    \includegraphics[width=0.23\textwidth]{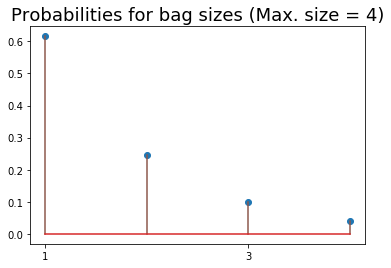}\hfill
    \includegraphics[width=0.23\textwidth]{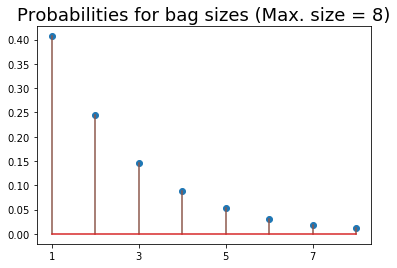}\hfill
    \includegraphics[width=0.23\textwidth]{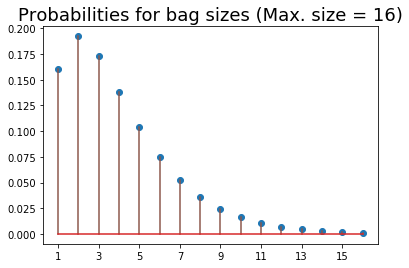}\hfill
    \includegraphics[width=0.23\textwidth]{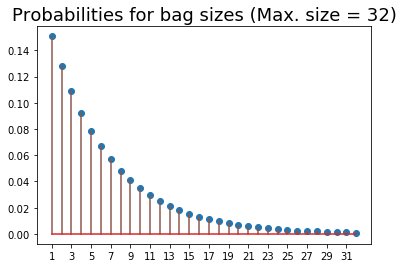}
    \caption{Multi-instance bag simulation: probabilities for sampling bag sizes for a given maximum bag size (from left to right: $4$, $8$, $16$, $32$).}
    \label{fig:bagsizes}
\end{figure}    

We generated a set of simulated exposures 
 from a fine uniform quantization of the three dimensional grid,
 corresponding to
 duration $\times$ distance $\times$ symptom onset.\footnote{
 The empirical distribution of duration, distance and onset can be estimated using the Exposure Notifications Privacy Analytics tool (see
 \url{https://implementers.lfph.io/analytics/}).
 Some summary statistics shared with us by the MITRE organization, that runs this service, confirms that the marginal distributions of each variable are close to uniform.
 } %
 For each point in this 3d grid, we sampled a label  from the Bernoulli distribution parameterized by the probability of infection using the 
 infection model described in \cref{sec:infModel}.
 
 After creating this "pool" of exposure events, we next assigned
 a random bag of size $k$ of these events to each user.
 The value $k$ is sampled from a truncated negative binomial distribution with parameters $(p,r)$, where the truncation parameter is the maximum bag size $b$.
 \Cref{fig:bagsizes} shows the probabilities of bag sizes for different settings of maximum bag size $b$. 
 Note that larger bag sizes make the multi-instance learning problem harder, because of the difficulty of "credit assignment".

Negative bags are composed of all negative exposures. We consider two scenarios for constructing positive bags: (i) each positive bag contains exactly one positive exposure and rest are negative exposures, (ii) each positive bag contains $N$ positive exposures, where $N$ is sampled uniformly from $\{1,2,3\}$. 
When there is only one positive in a bag, the learning problem is harder,
since it is like finding a "needle in a haystack".
(This corresponds to not knowing which of the people that you encountered actually infected you.)
When there are more positives in a bag, the learning problem is a little easier, since there are "multiple needles", any of which are sufficient to explain the overall positive infection.

To simulate censored exposures, we censor the positive exposures in each positive bag independently and identically with probability varying in the range $\{0.05, 0.1, 0.2, \ldots, 0.8\}$. We do not censor the negative exposures to prevent the bag sizes from changing too much from the control value.

\input{weaklearn_fig}

We use 80\% of the data for training, and 20\% for testing.
We fit the model using 1000 iterations
of (projected) stochastic gradient descent,
with a batch size of 100.
We then study the  performance (as measured by the area under the ROC curve, or AUC) as a function of the problem difficulty along two dimensions: (i) multi-instance learning (\ie, bag-level labels), 
and (ii) label noise (\ie, censored exposures). 
We compare the performance of the learned parameters with that of
two baselines.
The first corresponds to the oracle performance using the true probabilities coming from the simulator.
The second is a more realistic comparison,
namely the risk score configuration used by
the Swiss version of the
 GAEN app \citep{Salathe2020},
 whose parameters have been widely copied by many other health authorities.
This parameter configuration
uses 2 attenuation thresholds of 
$\bleThresh_1=53$ and $\bleThresh_2=60$.
The corresponding weights for the 3 bins are $\bleWeights_1=1.0$,
$\bleWeights_2=0.5$ and $\bleWeights_3=0.0$ (so exposures
in the third bin are effectively ignored).
The infectiousness weight is kept constant for all symptom onset values,
so is effectively ignored.

\subsubsection{Results}

We do 5 random trials, and report the mean and standard error of the results.
\Cref{fig:gridworld_auc} and \cref{fig:gridworld_auc_maxpos3} show the results for
two ways of  constructing positive bags: (i) each positive bag containing exactly one positive exposure, and (ii) each positive bag containing up to three positive exposures.
Not surprisingly, the oracle performance is highest.
However, we also see that the learned parameters outperform the Swiss baseline in all settings, often by a large margin.

In terms of the difficulty of the learning problem,
the general trend is that the AUC decreases with increasing bag size as expected. The AUC also decreases as the censoring probability is increased. The only exception is the case of fully observed exposures, when each positive bag can have up to three positive exposures, where the AUC increases in the beginning as the bag size is increased (up to bag size of $4$) and then starts decreasing again. This is expected since the learning algorithm gets more signal  to identify positive bags. This signal starts fading again as the bag size increases beyond $4$.  

\subsubsection{Robustness to model mismatch}
\label{sec:modelmismatch}

\begin{figure}
    \centering
    \begin{tabular}{cc}
    \includegraphics[width=2in]{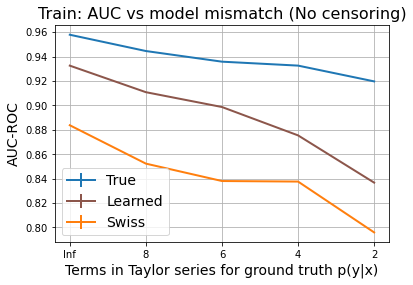}
        &
    \includegraphics[width=2in]{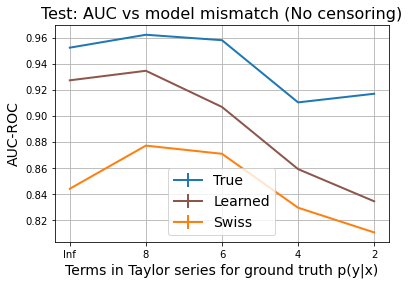}
\end{tabular}
    \caption{
    Learning performance (AUC) with increasing model mismatch for train and test set. We do not censor the exposures and the bag size is fixed to $4$. Each positive bag may contain up to three positive exposures (same setup as in Fig. \ref{fig:gridworld_auc_maxpos3}). The AUC worsens with increasing model mismatch but the learned scores still outperform the `Swiss' manual configuration.}
     \label{fig:model_mismatch}
\end{figure}

For a given risk score $r_n$, we take the functional form of the learned risk score model to be $p_l(y_n|x_n)=1-\exp(-\mu r_n)$ in all earlier experiments. This functional form matches the exponential dose response model used in our simulator where the probability of infection is given by $p_s(y_n|x_n)=1-\exp(-\lambda s_n)$ for hazard $s_n$ (note that $r_n$ and $s_n$ still have different functional forms and are based on different inputs). Here we simulate model mismatch in $p(y_n|x_n)$ and do preliminary experiments on how it might impact the performance of the learned risk score model.

In more detail,  we use $p_s(y_n|x_n)=1-f_t(-\mu s_n)$ for the dose response model in the simulator, where $f_t$ is the truncated Taylor series approximation of the exponential with $t$ terms.
This reflects the fact that our simulator may not be a very accurate of how \COVID is actually transmitted.
The functional form of learned model $p_l(y_n|x_n)$ is kept unchanged,
since that is required by the app.
We vary $t\in\{2,4,6,8\}$ and plot the AUC for the learned model in Figure \ref{fig:model_mismatch}. We fix the bag size to $4$ with each bag containing up to three positive exposures randomly sampled from $\{1,2,3\}$. As expected, the AUC gets worse as the model mismatch increases, however, the learned scores still outperform the manual `Swiss' configuration. We hypothesize that increasing the model capacity for computing the risk score $r_n$ (\eg, using a multilayer neural network or Gaussan Processes) can impart some robustness to model mismatch, but we leave as a direction for future work.

%% file: weaklearn_fig.tex
\begin{figure}
    \centering
    \begin{tabular}{ccc}
    \includegraphics[width=2in]{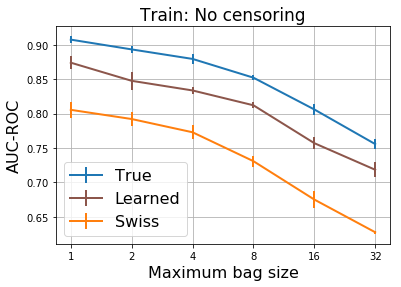}
        &
    \includegraphics[width=2in]{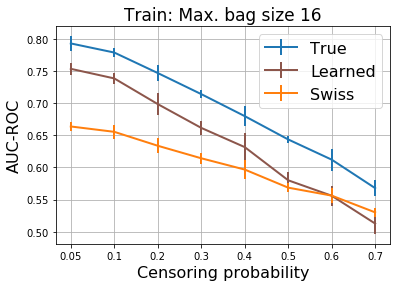}
    &
\includegraphics[width=2in]{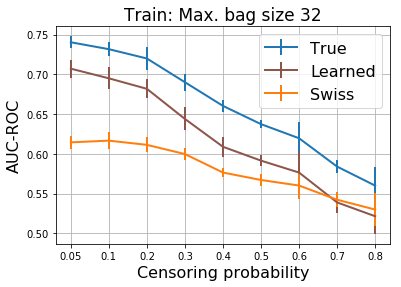}
\\
     \includegraphics[width=2in]{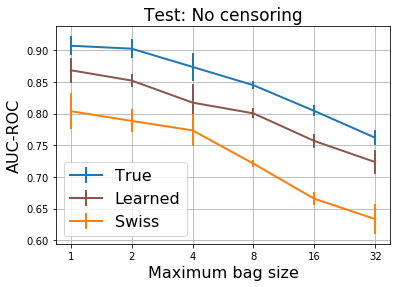}
        &
    \includegraphics[width=2in]{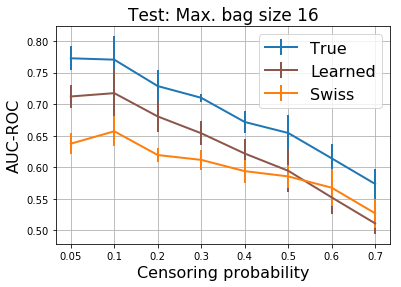}
    &
\includegraphics[width=2in]{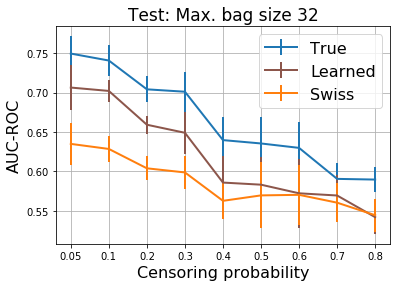}
\end{tabular}
    \caption{Learning performance (AUC) with increasing problem difficulty.
    Top row: training set.
    Bottom row: test set.
    Left column: No censoring. Performance vs bag size. 
    Middle column: Performance vs censoring probability, bag size 16.
    Right column: Performance vs censoring probability, bag size 32.
 Each positive bag contains \emph{exactly one} positive exposure.
 }
    \label{fig:gridworld_auc}
\end{figure}

\begin{figure}
    \centering
    \begin{tabular}{ccc}
    \includegraphics[width=2in]{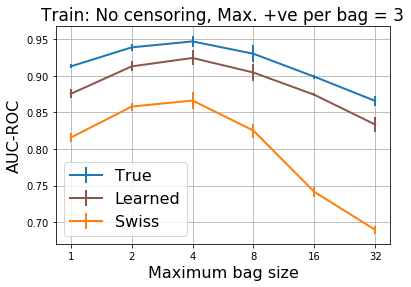}
        &
    \includegraphics[width=2in]{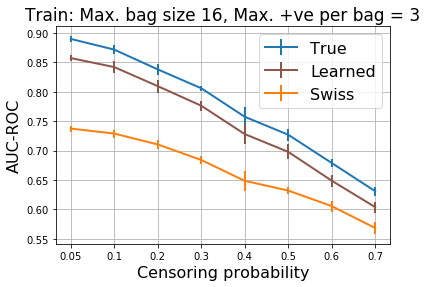}
    &
\includegraphics[width=2in]{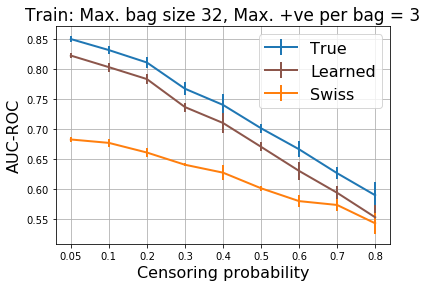}
\\
     \includegraphics[width=2in]{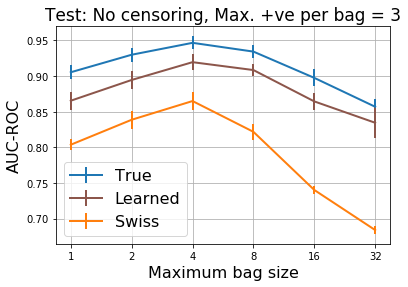}
        &
    \includegraphics[width=2in]{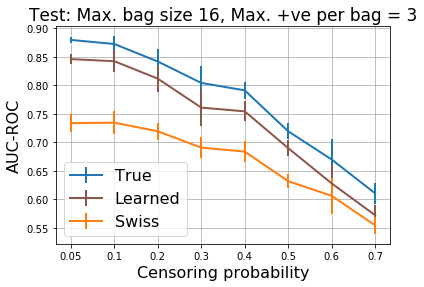}
    &
\includegraphics[width=2in]{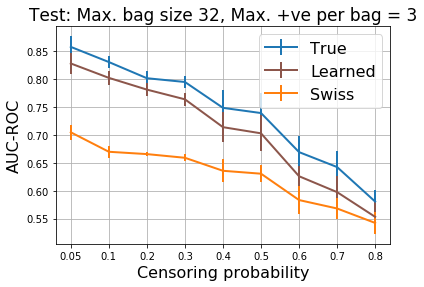}
\end{tabular}
    \caption{
    Same as \cref{fig:gridworld_auc} except now
     each positive bag may contain up to three positive exposures (number sampled uniformly from $\{1,2,3\}$.
 }
     \label{fig:gridworld_auc_maxpos3}
\end{figure}

%% file: related.tex
\section{Related work}
\label{sec:related}

Although there have been several papers which simulate the benefits of
using digital contact tracing apps such as GAEN
(e.g., \citep{Ferretti2020,Abueg2021}),
all of these papers assume the risk score model is known.
This is also true for papers that focus on inferring the individual
risk using graphical models of various forms 
(e.g., \citep{Cencetti2020,Herbrich2020}).
 
The only paper we are aware of that tries to learn the risk score
parameters from data is  \citep{Sattler2020}.
 They collect a small ground truth dataset of distance-attenuation pairs from 50 pairs of soldiers walking along a prespecified grid, whereas we use simulated data.
 However, they only focus on distance and duration, and ignore infectiousness of the index case.
 By contrast, our risk score model matches the form used in the GAEN app, so the resulting learned parameters could (in principle) be deployed "in the wild".
 
 In addition,  \citep{Sattler2020} uses standard supervised learning,
 whereas we consider the case of weakly supervised learning,
 in which the true infection outcome from each individual exposure event is not directly observed; instead users only get to see the label for their entire  "bag" of exposure events (which may also have false negatives due to censoring).
 Such weakly supervised learning is significantly harder from an ML point of view,
 but is also much more realistic of the kinds of techniques that would be needed in practice.

%% file: discussion.tex
\section{Discussion}

We have shown that it is possible to optimize the parameters of the risk score model for the Google/Apple \COVID app using machine learning methods,
even in the presence of considerable missing data and model mismatch.
The resulting risk score configuration resulted in much higher AUC scores than those produced by a widely used baseline.

\paragraph{Limitations}
The most important limitation of this paper is that this is a simulation study, and does not use real data (which was not available to the authors).

The second limitation is that
we have assumed a centralized, batch setting, in which a single dataset is collected, based
on an existing set of users, and then the parameters are learned, and broadcast back to new app users.
However, in a real deployment, the model would have to be learned online, as the data streams in, and the app parameters would need to be periodically updated. (This can also help if the data distribution is non-stationary,  due to new variants, or the adoption of vaccines.)

The third limitation is that we have not considered privacy issues.
The GAEN system was designed
from the ground-up to be privacy preserving,
and follows similar principles to 
the PACT (Private Automated Contact Tracing)
protocol (\url{https://pact.mit.edu/}).
In particular,
all the features are collected anonymously.
However, joining these exposure features $x$ 
with epidemiological outcomes $y$ in a central server
might require additional  protections,
such as the use of differential privacy
(see e.g., \citep{Abadi2016}),
which could reduce statistical performance.
(The privacy / performance tradeoff in DCT 
is discussed in more detail in \citep{Bengio2021}.)
We could avoid the use of a central server if we used
federated learning (see e.g.,
\citep{Kairouz2021}),
but this raises many additional challenges
from a practical and statistical standpoint.

The fourth limitation is more fundamental,
and stems from our finding that 
learning performance degrades when the recorded data does not contain all the 
 relevant information. This could be a problem in practical settings,
 where app usage may be rare, so that the cause of
 most positive test results remain "unexplained". Without enough data to learn from, any ML method is limited in its utility.
